\documentclass[12pt, letter]{article}
\usepackage[utf8]{inputenc}
\usepackage{hyperref}       
\usepackage{url}            
\usepackage{booktabs}       
\usepackage{amsfonts}       
\usepackage{nicefrac}       
\usepackage{microtype}      
\usepackage{amsmath}
\usepackage{xcolor}
\usepackage[margin = 1in]{geometry}
\usepackage[inline]{enumitem}
\usepackage{float}
\usepackage{graphicx}
\usepackage{subcaption}
\usepackage{amsthm}
\usepackage{amsmath}
\usepackage{bbm}
\usepackage{verbatim}

\usepackage{algorithm}
\usepackage{algorithmic}
\usepackage{bm}
\usepackage{caption}
\usepackage{tabularx}

\newcommand{\XCal}{\mathcal{X}}
\newcommand{\YCal}{\mathcal{Y}}
\newcommand{\ZCal}{\mathcal{Z}}

\newcommand{\LCal}{\mathcal{L}}
\newcommand{\E}{\mathbb{E}}

\newcommand{\Cov}{\mathrm{Cov}}
\newcommand{\Var}{\mathrm{Var}}

\title{Meta-learning with GANs for anomaly detection, with deployment in high-speed rail inspection system}
\author{Haoyang Cao\footnote{Centre de Math\'ematiques Appliqu\'ees, \'Ecole Polytechnique, Route de Saclay, 91128, Palaiseau Cedex, France, haoyang.cao@polytechnique.edu}\and Xin Guo\footnote{Department of IEOR, University of California, Berkeley, Berkeley, California, 94709, United States, xinguo@berkeley.edu} \and Guan Wang\footnote{Tsinghua-Berkeley Shenzhen Institute, Shenzhen, P. R. China, wangguan17@mails.tsinghua.edu.cn}}
\date{\today}

\begin{document}
\maketitle
\section*{Abstract}
Anomaly detection has been an active research area with a wide range of potential applications. Key challenges for anomaly detection in the AI era with big data include lack of prior knowledge of potential anomaly types, highly complex and noisy background in input data, scarce abnormal samples, and imbalanced training dataset. In this work, we propose a meta-learning framework for anomaly detection to deal with these issues. Within this framework, we incorporate the idea of generative adversarial networks (GANs) with appropriate choices of loss functions including structural similarity index measure (SSIM). Experiments with limited labeled data for high-speed rail inspection demonstrate that our meta-learning framework is sharp and robust in identifying anomalies. Our framework has been deployed  in five high-speed railways of China since 2021: it has reduced more than 99.7\% workload and saved 96.7\% inspection time. 

\section{Introduction}
Anomaly detection has been an active research area with a wide range of potential applications such as fraud detection in health insurance, cyber attacks, and military surveillance \cite{chandola2009anomaly}.
The goal of anomaly detection is to identify abnormal input data that does not conform to a given data description.
\paragraph{Motivating example of anomaly detection.}
High-speed rail inspection is one example of anomaly detection that embodies the key challenges in the AI era.
In a high-speed rail system, identification and removal of foreign abnormal objects is critical for the safety of railway operation. However, there are limited number of inspectors in contrast to hundreds of thousands of miles of railway to inspect.
Current practice in China, for instance, is first to take photos of the track and its vicinity using a camera on a moving train, and then to manually inspect potential abnormal objects from the photos. This manual inspection process is tedious and prone to human errors: for each rail line there are over tens of thousands of images taken each week; and for each image it takes from 30 seconds to a minute for each inspector to check and record. This amounts to hundreds of working hours for each line, even with the aid of industrial high-definition cameras to digitize track status.

Speedy and accurate identification of anomaly for high-speed rail is a formidable task, for several reasons. First, climatic and geographic changes lead to different distributions of images collected from the rail lines, and the background of images are highly noisy, diverse, and complex.
 Second, all kinds of foreign objects can fall onto or near the track. Therefore, it is infeasible to know {\it a priori} all anomaly types. Third, 
 among all potential anomaly types, those captured by inspection photos are extremely limited and such instances are negligibly infrequent. Thus data are highly imbalanced (see Columns (A) and (B) in Table \ref{subtab: 100}).
 Finally, labeled data are extremely limited, as labeling hundreds of thousands of images is labor intensive.

\paragraph{Main challenges.}
The example of high-speed rail anomaly detection highlights some of main challenges for the most formidable anomaly detection problems, including \begin{itemize}
\item Highly diverse, complex and noisy background in input data;
 \item No prior knowledge of anomaly types;
 \item Exceedingly imbalanced training dataset, with very limited or no abnormal sample data.
 \end{itemize}

\paragraph{Existing approaches for anomaly detection.}
 Early works on anomaly detection \cite{tan2005introduction} adopt classification techniques, including neural networks, Bayesian networks, support vector machine, and nearest neighbor \cite{chandola2009anomaly}. 
These traditional supervised-learning methods face the overfitting issue when abnormal samples are limited and the training dataset is imbalanced.
Alternatively, unsupervised learning approaches such as clustering-based anomaly detection \cite{tan2005introduction} have been developed. These methodologies, however, reply on the crucial assumption that normal data lie close to their closest cluster centroid whereas anomalies are far away from those clusters. Moreover, techniques such as k-means clustering can be negatively impacted by outliers and are not suitable for identifying clusters with irregular shapes. 
Additionally, semi-supervised deep learning techniques have been proposed for anomaly detection, including long-short term memory \cite{1997Long}, convolutional neural network \cite{NIPS2012_c399862d}, autoencoder \cite{hinton2006reducing}, deep neural network \cite{2012Deep}, spatial transformer network \cite{0Spatial}, adversarial autoencoder \cite{2015Adversarial}, variational autoencoder \cite{2016An}, and
generative adversarial networks (GANs) (\cite{cheng2020adgan,luer2019anomaly,schlegl2019f}. These models learn patterns and features of the normal data distribution, and produce outstanding performance for detecting anomalies. 
 Additionally, {\em model-agnostic meta-learning} framework has been adopted to solve multiple anomaly detection tasks (\cite{lu2020few,wu2021learning,9291099}). However, all these techniques require refined and simple backgrounds against which abnormal samples are distinctive. 

\paragraph{Previous works on railway track inspection.}
In \cite{giben2015material}, end-to-end deep convolutional neural networks (DCNN) are adopted to inspect fatigue cracks of timber using material classification and semantic segmentation. The work of \cite{gibert2015robust} adopts the histogram of oriented gradients features and a combination of linear SVM classifiers to detect fasteners and ties of the track. Similar work has been done in \cite{feng2013automatic} to detect partially worn and completely missing fasteners using probabilistic topic model. \cite{kang2018deep} designs an R-CNN network for the insulator inspection. 
All techniques from these works for anomaly detection belong to the category of supervised learning and require abundant labeled data with known anomaly types. 

\paragraph {Our approach.}
The goal of our study is to develop an anomaly detection framework that requires little-to-no labeled training data which may be highly imbalanced and to produce accurate detection regardless of the image backgrounds.

\begin{itemize}
 \item To deal with different distributions of samples images, we propose and develop a {\em meta-learning framework} to solve {\em multiple tasks} of anomaly detection. 
 
\item In the meta-learning framework, the meta objective is designed to restore images and to correct any possible anomalies, and the individual tasks are binary classifications for collected images. In the meta objective, we incorporate the idea of generative adversarial networks (GANs) \cite{Goodfellow2014Generative} to ensure a credible reconstruction of target images from input; we also adopt the structural similarity index measure (SSIM) to enhance the identification of the anomalies. This addition of GANs and SSIM resolves the issue of unlabeled data.
 
 \item We design neural network architectures that are tailored for this meta-learning framework.
\end{itemize}
\paragraph{Our results.}
We first experiment our meta-learning framework with limited labeled data to test its accuracy and robustness and eventually deploy it within the high-speed rail inspection system with no labeled data.

The experiment demonstrates superb performance of this meta-learning framework: with both limited labeled data and limited known anomaly samples, it rarely misses any anomalies that were previously detected by human inspectors and it is robust with respect to the reduction in the number of labeled samples. Moreover, it can detect several folds more anomalies than human inspectors. 
 
Our framework has been deployed  in five high-speed railways of China since 2021. Without any prior labeled data, it has detected several times more new anomalies than human inspectors; it has  reduced the workload of inspecting millions of images to simply double-checking hundreds of identified images,  a workload reduction of over
99.7\%; and  it takes less than two days for anomaly detection, instead of several months for purely manual inspection, 96.7\% inspection time saved.

\section{Technical Components for High-Speed Rail Anomaly Detection}
\subsection{Review}
\paragraph{Meta learning.}
As reviewed in \cite{hospedales2020meta},
meta-learning is a broad learning paradigm.
Combined with deep neural networks and reinforcement learning techniques, meta learning has gained substantial attention in model-agnostic learning \cite{finn2018probabilistic}, online learning \cite{finn2019online}, memory-augmented neural networks \cite{santoro2016meta} and latent-embedded optimization \cite{rusu2018metalearning}.

In contrast to conventional machine learning where {the learning tasks are often solved independently}, meta-learning leverages the experience of multiple learning episodes. Under a multi-task scenario, a meta-learning framework typically specifies a set of task-independent features among all tasks. On top of these tasks, a different optimization problem is employed with respect to these task-independent features. This optimization is often referred to as the {\em meta objective} and is for improving the learning outcome of these individual tasks. 

A typical meta-learning framework across $K\in\mathbb{N}$ various tasks can be seen as the following bi-level optimization problem,
\begin{equation}
 \label{eq: meta-meta}
 \min_{\omega\in\Omega}\sum_{i=1}^K\LCal_{{\rm meta}}(\theta_i^*(\omega),\omega),
\end{equation}
where for $i=1,\dots,K$, $\theta_i^*(\omega)$ is defined as
\begin{equation}\label{eq: meta-ind}
 \theta_i^*(\omega)\in\arg\min_{\theta\in\Theta_i}\LCal_{i}(\theta,\omega),\quad \forall\omega\in\Omega.
\end{equation}
Here, the outer optimization Equation \eqref{eq: meta-meta} with loss function $\LCal_{{\rm meta}}$ is seen as the meta objective, and the inner optimization Equation \eqref{eq: meta-ind} with loss function $\LCal_{i}$ is seen as the task-specific objective. The set $\Omega\subset\mathbb{R}^k$ contains all possible task-independent parameter $\omega$ shared across all $K$ tasks. $\Theta_i\subset\mathbb{R}^{d_i}$ denotes the set of task $i$-specific parameters $\theta$, for $i=1,\dots,K$. Loss functions $\LCal_{{\rm meta}}$ and $\LCal_i$'s are to optimize task-independent parameter $\omega$ and task-specific parameters $\theta$, respectively.

Proposing meta-learning framework for anomaly detection is to overcome the obstacle in the traditional classification formulation with unlabeled data.
 
\paragraph{GANs.}
Generative adversarial networks (GANs), introduced in \cite{Goodfellow2014Generative}, belong to the class of generative models. 
Since the inception, GANs have enjoyed tremendous empirical successes in high resolution image generation \cite{denton2015deep,radford2015unsupervised}, 
text-to-image synthesis \cite{reed2016generative}, video generation \cite{vondrick2016generating}, semantic segmentation \cite{luc2016semantic}, and abstract reasoning diagram generation \cite{ghosh2016contextual}. 

The key idea behind GANs is to interpret the process of generative modeling as a competing game between two neural networks: a generator neural network $G_{\omega_G}$ with parameters $\omega_G$ and a discriminator neural network $D_{\omega_D}$ with parameters $\omega_D$. The generator network $G_{\omega_G}$ attempts to fool the discriminator network by mapping random noise $Z$ from a latent space $\ZCal$ into the sample space $\XCal$, while the discriminator network $D_{\omega_D}$ tries to identify whether an input sample is faked or true. A vanilla formulation in \cite{Goodfellow2014Generative} is a minimax problem,
\begin{equation}\label{eq: GANs-def}\min_{\omega_G}\max_{\omega_D} \E[\log(D_{\omega_D}(X))+\log(1-D_{\omega_D}(G_{\omega_G}(Z)))].\end{equation}
Other frameworks of GANs include f-GAN \cite{nowozin2016f}, Wasserstein GAN \cite{arjovsky2017wasserstein}, relaxed-Wasserstein GAN \cite{guo2021relaxed}, and a survey paper \cite{cao2021generative}. 

Incorporating GANs into the meta objective for anomaly detection is to ensure a credible production of the target image from the input image.

\paragraph{SSIM}
Structural similarity index measure is a perception-based model widely used in the field of image reconstruction \cite{2020Reconstruction}, object detection \cite{2019BASNet}, super-resolution \cite{2014Single}, and image deblurring \cite{2016A}. 

The key idea behind SSIM is to perceive image degradation as structural changes in luminanice and contrast. Structural similarity comparison is based on the notion that pixels have strong inter-dependence, especially when they are spatially close and these dependencies carry important visual information about the structure of the objects \cite{2004Image}. Luminance comparison comes from the experience that image distortions tend to be less visible in bright regions, whereas contrast comparison is built on the intuition that distortions become less visible where there is significant amount of activity or texture in the image. 

For two sample images $Y,Y'\in \mathbb{R}^{w\times h\times 3}$ of width $w\in\mathbb{N}$ and height $h\in\mathbb{N}$, the SSIM loss \cite{wang2004image} is given by
\begin{equation}\label{eq: SSIM-def-0}\begin{aligned}L_{\rm{SSIM}}^{\alpha,\beta,\gamma}(Y,Y')&=l(Y,Y')^\alpha\\&\cdot c(Y, Y')^\beta\cdot s(Y,Y')^\gamma,\end{aligned}\end{equation}
where $\alpha, \beta, \gamma$ are constants and the functions $l(Y,Y')$, $c(Y,Y')$ and $s(Y,Y')$ represent image illumination comparison, contrast comparison, and structural similarity comparison between $Y$ and $Y'$, respectively, and
\begin{equation}\label{eq: SSIM-def}
\begin{aligned}
 &l(Y,Y')=\frac{\sum_{i=1}^{w}\sum_{j=1}^{h}\sum_{k=1}^{3}\frac{2\E_Y[Y_{ijk}]\E_{Y'}[Y'_{ijk}]+c_1}{\E_Y[Y_{ijk}]^2+\E_{Y'}[Y'_{ijk}]^2+c_1}}{w\times h\times 3},\\
 &c(Y,Y')=\frac{\sum_{i=1}^{w}\sum_{j=1}^{h}\sum_{k=1}^{3}\frac{2\Cov_{Y,Y'}(Y_{ijk},Y'_{ijk})+c_2}{\Var_Y(Y_{ijk})+\Var_{Y'}(Y'_{ijk})+c_2}}{w\times h\times 3},\\
 &s(Y,Y')=\frac{\sum_{i=1}^{w}\sum_{j=1}^{h}\sum_{k=1}^{3}\frac{\Cov_{Y,Y'}(Y_{ijk},Y'_{ijk})+c_3}{\Var_Y(Y_{ijk})\Var_{Y'}(Y'_{ijk})+c_3}}{w\times h\times 3},
\end{aligned}
\end{equation}
with $c_1$, $c_2$ and $c_3$ some constants.

Including SSIM loss into the meta-learning framework for anomaly detection is 
to emphasize the contrast between input images (especially those containing abnormal areas) and their reconstructed version, and hence highlight the anomaly correction step in the meta objective.

\subsection{Meta-learning framework for anomaly detection}
\label{subsec: framework}

\paragraph{Notation.}
Since the sample dataset consists of image data, we use $\XCal\subset\mathbb{R}^{w\times h\times3}$ to denote the sample space, where $w$ and $h$ are the width and the height of sample images. For any sample image $X\in\XCal$, for instance an image of a railway track segment, we pair it with a target image $Y\in\XCal$ which is a clean version of $X$ without any anomaly, and denote the pair as $(X,Y)$. 

\paragraph{Meta objective.}
The first building block is a generative model to reconstruct an image of $Y$ via a mapping $G:\XCal\times\ZCal\to\XCal$, where $\ZCal$ is a latent space. 
That is, $G$ takes a sample image $X\in\XCal$ and a latent random variable $Z\in\ZCal$ as inputs, and produces a corrected version of $X$, namely $G(X,Z)$, so that the generated pair $(X,G(X,Z))$ resembles the pair $(X,Y)$ from the training dataset. 

The second building block is to choose appropriate measures between the generated images and the original ones, including
\begin{enumerate}
 \item the divergence between the distribution of $(X,G(X,Z))$ and that of $(X,Y)$;
 \item the perceptual difference between $G(X,Z)$ and $Y$.
\end{enumerate}

To handle the distributional divergence, we will adopt GANs \cite{Goodfellow2014Generative} and introduce discriminator $D:\XCal\times\XCal\to[0,1]$ to estimate the Jensen--Shannon divergence between the joint distribution of $(X,G(X,Z))$ and that of $(X,Y)$. 
To measure the distance between images $G(X,Z)$ and $Y$, we will adopt both the $L_2$ loss and structural similarity loss (SSIM). Combined, the meta objective of anomaly detection is formulated as the following minimax model,
\begin{equation}
 \label{eq: meta-out}
 \min_{\omega_G}\max_{\omega_D}\mathcal L_{\rm{meta}}(G_{\omega_G},D_{\omega_D}).
\end{equation}
Here, $G_{\omega_G}:\XCal\times\ZCal\to\XCal$ and $D_{\omega_D}:\XCal\times\XCal\to[0,1]$ denote two parametrized neural networks approximating $G$ and $D$, respectively. Their corresponding parameters, $\omega_G$ and $\omega_D$, constitute the task-independent parameter $\omega=(\omega_G,\omega_D)\in\Omega\subset\mathbb{R}^k$ of the meta objective; here the dimension $k$ denotes the number of parameters to train in both networks $G_{\omega_G}$ and $D_{\omega_D}$. Moreover, the objective function $\mathcal L_{{\rm meta}}$ is modified from GANs' objective and defined as
\begin{equation}
 \label{eq: meta-obj}
 \begin{aligned}
 \mathcal{L}_{\rm{meta}}(G_{\omega_G},D_{\omega_D})&=V_{\rm{GAN}}(G_{\omega_G},D_{\omega_D})\\
 &+w_1V_{L_2}(G_{\omega_G})-w_2V_{\rm{SSIM}}(G_{\omega_G}).
 \end{aligned}
\end{equation}
Here, $V_{\rm{GAN}}$ takes the form of the loss function for vanilla GANs \eqref{eq: GANs-def} such that
\[\begin{aligned}V_{\rm{GAN}}(G_{\omega_G},D_{\omega_D})&=\E_{X,Y}\log D_{\omega_D}(X,Y)\\
&+\E_{X,Z}\log\left[1-D_{\omega_D}\left(X,G_{\omega_G}(X,Z)\right)\right],\end{aligned}\] 
$V_{L_2}$ and $V_{\rm{SSIM}}$ are given by
\[V_{L_2}(G_{\omega_G})=\E\left\|Y-G_{\omega_G}(X,Z)\right\|^2,\]
and 
\[\begin{aligned}V_{\rm{SSIM}}(G_{\omega_G})&=L_{\rm{SSIM}}^{1,1,1}(Y,G_{\omega_G}(X,Z)=l(Y,G_{\omega_G}(X,Z))\\&\cdot c(Y,G_{\omega_G}(X,Z))\cdot s(Y,G_{\omega_G}(X,Z)),\end{aligned}\]
where functions $l(Y,Y')$, $c(Y,Y')$ and $s(Y,Y')$ represent image illumination comparison, contrast comparison, and structural similarity comparison of random samples $Y,Y'\in\XCal$, respectively, according to Equation \eqref{eq: SSIM-def}.

Having $V_{\rm{GAN}}$ as part of the meta objective is to ensure a credible reproduction of the input image. It also differentiates our meta-learning framework for anomaly detection from conventional meta-learning models: instead of a pure minimization problem such as \eqref{eq: meta-meta}, our meta objective \eqref{eq: meta-out} solves for a saddle point $\omega^*=(\omega_G^*,\omega_D^*)$ of a minimax problem because of the composition of Equation \eqref{eq: meta-obj}; another noticeable difference is that our meta objective \eqref{eq: meta-out} does not depend on the results of individual tasks and it is due to the generative nature of $G_{\omega_G}$. In addition, to strengthen the correction ability of $G_{\omega_G}$, $V_{\rm{GAN}}$ is enhanced on the generator side by adding $L_2$ loss \cite{pathak2016context} and structural similarity (SSIM) loss \cite{wang2004image}.


\paragraph{Task-specific objective.}
For anomaly detection problem of complex backgrounds, we disintegrate it into multiple binary classification problems so that within each task, the input images are assume to follow the same distribution.

Previously in the meta objective, the goal is for the network $G_{\omega_G}$ to restore input images and correct any possible anomalies. However, $G_{\omega_G}$ itself will not determine the anomaly of the original image. 
After solving the meta objective and fix a task-independent parameter $\omega=(\omega_G,\omega_D)$ from the minimax model, we now formulate the anomaly detection as a binary classification problem. 

In particular, for any given task $i=1,\dots,K$ and the corresponding samples $X^i$, first its corrected sample $\hat X^i(\omega)=G_{\omega_G}(X^i,Z)$ is construct. Then, the SSIM loss is computed for $(X^i,\hat X^i(\omega))$ and with a predetermined threshold value $\mu_0>0$, one can assign a corresponding label $\bm{1}_{X^i}\{{\rm normal}\}$ and extract the appropriate classifier input $f(X^i,\hat X^i(\omega))$ in a feature space $\YCal$. A detailed labelling and feature extraction procedure can be found in Section \ref{sec: exp}. Let $\theta_i\in\Omega_i\subset\mathbb{R}^{d_i}$ be the individual task parameter for the binary classifier $p_{\theta_i}:\YCal\to[0,1]$ for task $i$. Here, the dimension $d_i$ specifies the number of parameters to train in the classifier network, and $p_{\theta_i}\left(f(X^i,\hat X^i(\omega))\right)$ is the probability of $X^i$ being a normal image. Then the task-specific loss is given by cross-entropy value,
\begin{equation}
 \label{eq: meta-in}
 \begin{aligned}
 \mathcal{L}_i(\theta_i,\omega)=&-\E_{X^i,Z}\biggl[\bm{1}_{X^i}\{\mathrm{normal}\}p_{\theta_i}\left(f(X^i,\hat X^i(\omega))\right)\\
 &+\left(1-\bm{1}_{X^i}\{\mathrm{normal}\}\right)\left[1-p_{\theta_i}\left(f(X^i,\hat X^i(\omega))\right)\right]\biggl],
 \end{aligned}
\end{equation}
and $\theta_i(\omega)\in\arg\min_{\theta_i}\mathcal{L}_i(\theta_i,\omega)$.

\section{Network architectures for the meta-learning framework}\label{nn}

Our meta-learning framework involves designing appropriate neural networks for both the minimax problem \eqref{eq: meta-out} as the meta objective and the classification problem \eqref{eq: meta-in} as the task-specific objective. 

\begin{figure}[!ht]
 \centering
 \includegraphics[width=\textwidth]{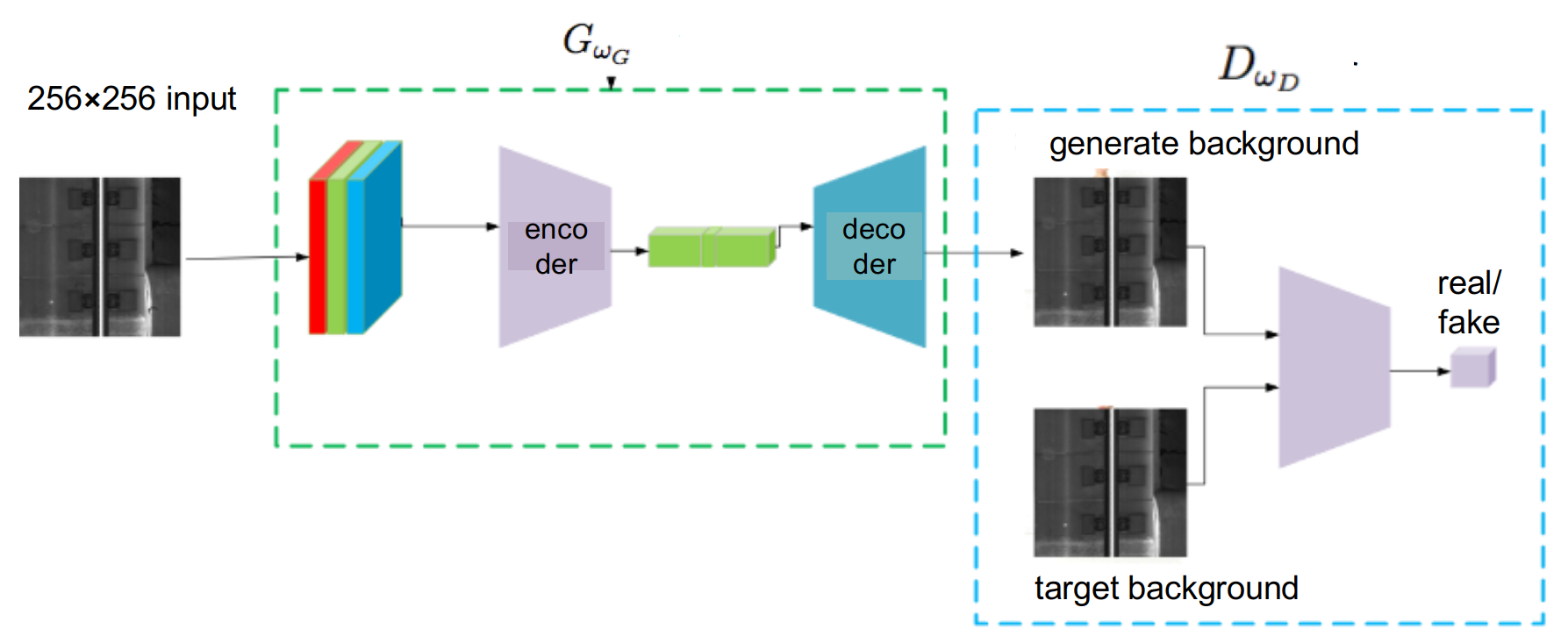}
 \caption{Network Structure for the Meta Objective}
 \label{fig:6}
 \end{figure}

\begin{figure}[!ht]
 \centering
 \includegraphics[width=\textwidth]{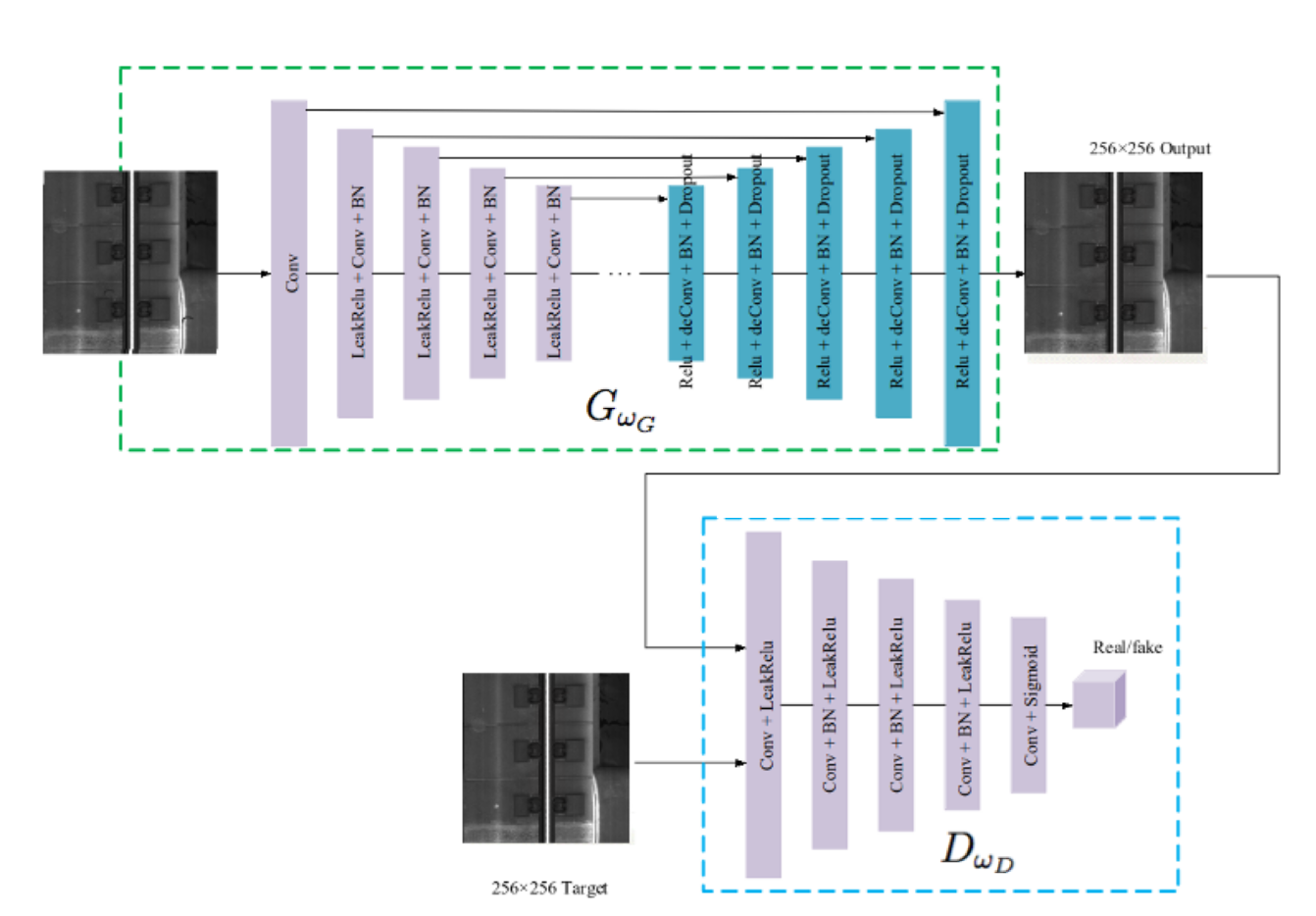}
 \caption{Details of Network Structure for the Meta Objective}
 \label{fig:7}
 \end{figure}

For the minimax problem \eqref{eq: meta-out}, the network structure includes two neural networks $G_{\omega_G}$ and $D_{\omega_D}$, as shown in Figure \ref{fig:6}.
The generator network $G_{\omega_G}$ consists of an encoder, a decoder, and a direct link called 
U-NET structure \cite{ronneberger2015u} between them to enhance the information passage. {More specifically, there are $2n$ layers (for instance, in this paper $n=8$) in the network $G_{\omega_G}$ with the encoder and the decoder each containing $n$ layers. For $i=1,\dots, n$, the output of the $i$-th layer in the encoder is passed to both the $(i+1)$-th encoder layer and the $(n-i)$-th decoder layer.} {The decoder then integrates this extra information from the encoder into the usual forward propagation through the decoder layers and generates normal images by removing possible anomalies.} {The discriminator network $D_{\omega_D}$ is composed of a 5-layer convolution network \cite{radford2015unsupervised}, with its details shown in Figure \ref{fig:7}.}
To better focus on local images, the network $D_{\omega_D}$ uses the PatchGAN architecture \cite{li2016precomputed}, which has been proved effective in the classification of local classifiers instead of global classifiers. PatchGAN also reduces the number of parameters, thereby improving the training efficiency.

For the task-specific objective \eqref{eq: meta-in}, the network adopted is the convolutional neural network (CNN) \cite{LeCun1999ObjectRW} and consists of the feature extractor and the decoder. The feature extractor has several convolutional layers and pooling layers. It captures basic features such as lines and corners in the first few layers and extracts more advanced features such as indicators of anomalies in the later layers. The extracted features are then fed into the decoder part. The decoder is a set of fully connected layers and uses extracted features to predict the target variable. In anomaly detection, the probability of an input image being anomaly is the target variable.
We adopt a shallow CNN model with only four layers: two convolution layers and two fully-connected layers. To prevent overfitting, we add dropout layers \cite{JMLR:v15:srivastava14a} in the decoder part and the $L^2$ regularization for the kernel of the fully connected layers.

\section{Experiments on high-speed rail detection dataset}\label{sec: exp}
We apply our meta-learning framework to the railway detection dataset. The dataset contains nearly a million images collected from  five high-speed rail lines in China (i.e., the number of individual tasks $K=5$). 
These images are not labeled and abnormal images can only be confirmed after manual verification from the detected samples. Foreign objects refer to objects that do not belong to the rail track itself. The most common ones are nails broken from the rail, stones, various types of garbage, and dead animals. 

Columns (A) and (B) in Table \ref{subtab: 100} contain basic information of these five rail lines, where the {\em Total Number of Images} refers to the total number of images in each rail line, and the {\em Official Anomaly Record} refers to the number of anomaly cases recorded by railway operation staff through regular inspections.

\begin{figure}[!ht]
 \begin{subfigure}[t]{0.3\textwidth}
    \includegraphics[width=\textwidth]{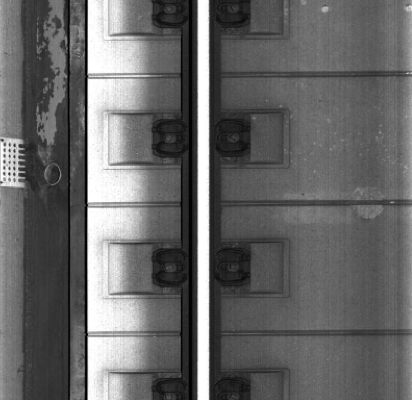}
    \caption{Original Image}
    \label{subfig: raw}
 \end{subfigure}
 \hspace{12pt}
 \begin{subfigure}[t]{0.3\textwidth}
    \includegraphics[width=\textwidth]{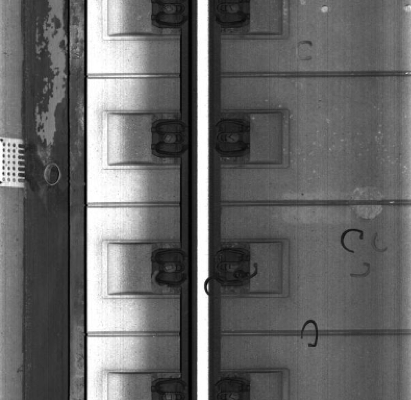}
    \caption{Synthesized Image}
    \label{subfig: synth}
 \end{subfigure}
 \hspace{12pt}
 \begin{subfigure}[t]{0.3\textwidth}
    \includegraphics[width=\textwidth]{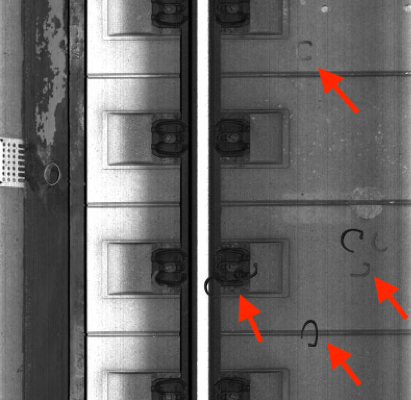}
    \caption{Marked Image}
    \label{subfig: marked}
 \end{subfigure}
 \caption{Example of Synthesized Abnormal Sample}
 \label{fig: preproc}
\end{figure}

\paragraph{Data pre-processing.}
For data pre-processing, we synthesize abnormal samples by pasting images of commonly seen foreign objects onto images originally collected from the rail lines. If sizes of foreign objects are disproportionately small compared with the original images, we group two to fifteen such foreign objects when pasting.
As a result, the complete set of training data $\XCal\subset\mathbb{R}^{256\times256\times3}$ includes the original images as well as these newly synthesized abnormal samples, both with $256\times256$ pixels. In the context of Section \ref{subsec: framework}, take any $X\in\XCal$, if $X$ is one of the original images then the corresponding $Y$ should be itself, that is, $Y=X$; if $X$ is one of the synthesized abnormal images, then the corresponding $Y$ is its original image. Figure \ref{fig: preproc} is an example of the synthesized abnormal samples during data pre-processing. Figure \ref{subfig: raw} is the original image captured from a test line and Figure \ref{subfig: synth} is the corresponding synthesized abnormal sample after pasting abnormal objects (e.g. nails) onto the original image; the abnormal regions on the synthesized image are marked by red arrows on Figure \ref{subfig: marked}.

\paragraph{Algorithm.}
Our Algorithm \ref{alg:1} of anomaly detection for railway inspection is divided into two steps: the meta-learning step and the line-specific anomaly detection step. 

In step one (the meta-learning step), for each rail line, we train the model described in Equations \eqref{eq: meta-out} and \eqref{eq: meta-in} through the pre-processed data and find the optimal parameters $\omega_G$ and $\omega_D$ for two neural networks \(G_{\omega_G}\) and \(D_{\omega_D} \) respectively. We use such optimal parameters as the meta knowledge for the anomaly reconstruction in each rail line. 

In step two (the line-specific anomaly detection step), for each task-specific problem, we divide the input image $X$ and reconstructed image $\hat X=G_{\omega_G}(X,Z)$ into $16\times16$ blocks $\left(X_{kl},\hat X_{kl}\right)$ for $k,l=1,\dots,16$, and compute the SSIM loss for each block pair $\left(X_{kl},\hat X_{kl}\right)$. We pick three block pairs $\left(X_{(j)},\hat X_{(j)}\right)$ for $j=1,2,3$ with the three lowest SSIM values, and compute the difference $\Delta X_{(j)}=X^i_{(j)}-\hat X_{(j)}$. For experiments with limited labeled data, the label for $X$ is given. In the deployment stage without labeled data, $X$ is labeled as {\em abnormal} if the lowest block SSIM loss is less than a predetermined threshold $\mu_0>0$. As described in Equation \eqref{eq: meta-in}, a binary classifier is trained with input $\left(\Delta X_{(1)}, \Delta X_{(2)},\Delta X_{(3)}\right)$.

\paragraph{Computing resources.} We utilize a workstation with a single Nvidia GeForce GTX 1080 GPU, an Intel Core i7-7700K CPU and 16GB memory for all experiments.

\begin{algorithm}[!ht]
	\renewcommand{\algorithmicrequire}{\textbf{Input:}}
	\renewcommand{\algorithmicensure}{\textbf{Output:}}
	\caption{Meta-Learning Algorithm for Anomaly Detection}
	\label{alg:1}
	\begin{algorithmic}[1]
	\footnotesize{
		\REQUIRE Hyper-parameters learning rate $\alpha$, $\beta$, $\gamma$\\ Initialize the network parameter $\omega=(\omega_G,\omega_D)$ of $G_{\omega_G},D_{\omega_D} $
		\STATE
		\STATE \*STEP ONE: META-LEARNING STEP
		\WHILE{not converged (or epoch \textless 300 )} 
		\STATE Sample a batch of lines $\{{L}_{i}\}^{N}_{i=1}$
		\FOR{each ${L}_{i}$}
		\STATE Construct task ${T}_{i}$ including training set, validation set for each line ${T}_{i} = ({D}_{i}^{tr},{D}_{i}^{val})$ from ${L}_{i}$
		\STATE Evaluate the gradient for $\omega_G$, $\bigtriangledown_{\omega_G}{L}_{T_i}(\omega;{D}_{i}^{tr})$, where
		\STATE ${L}_{T_i}(\omega;{D}_{i}^{tr})=\sum_{(X_j,Y_j)\in{{D}_{i}^{tr}}}{\mathcal L}_{\rm{meta}}(G_{\omega_G},D_{\omega_D})=\sum_{(X_j,Y_j)\in{{D}_{i}^{tr}}}[V_{\rm{GAN}}(G_{\omega_G},D_{\omega_D})+w_1V_{L_2}(G_{\omega_G})-w_2V_{\rm{SSIM}}(G_{\omega_G})]$
		\STATE Update $\omega_G$: $\omega_{G}\leftarrow \omega_G - \alpha \bigtriangledown_{\omega_G}{L}_{T_i}(\omega_G,\omega_D;{D}_{i}^{tr})$
		\ENDFOR
		\STATE Evaluate the gradient for $\omega_D$, 
		$\bigtriangledown_{\omega_D}{L}_{T_i}(\omega;{D}_{i}^{val})$, where
		\STATE ${L}_{T_i}(\omega;{D}_{i}^{val})=\sum_{(X_j,Y_j)\in{{D}_{i}^{val}}}{\mathcal L}_{\rm{meta}}(G_{\omega_{G}},D_{\omega_D})=\sum_{(X_j,Y_j)\in{{D}_{i}^{val}}}[V_{\rm{GAN}}(G_{\omega_G},D_{\omega_D})+w_1V_{L_2}(G_{\omega_G})-w_2V_{\rm{SSIM}}(G_{\omega_G})]$
		\STATE Update $\omega_D$: $\omega_{D}\leftarrow\omega_D - \beta \sum_{i=1}^{N} \bigtriangledown_{\omega_D}{L}_{T_i}(\omega;{D}_{i}^{val})$
		\ENDWHILE
		\STATE
		\STATE \*STEP TWO: LINE-SPECIFIC ANOMALY DETECTION STEP
		\FOR{each line ${L}_{i}$}
		\STATE Construct meta-test set ${D}_{i}^{test}$ for line $L_i$ and update $\omega_G$ again: $\omega_{G_i}\leftarrow \omega_G - \alpha \bigtriangledown_{\omega_G}{L}_{T_i}(\omega;{D}_{i}^{test})$, where
		\STATE ${L}_{T_i}(\omega;{D}_{i}^{test})=\sum_{(X_j,Y_j)\in{{D}_{i}^{test}}}{\mathcal L}_{\rm{meta}}(G_{\omega_G},D_{\omega_D})=\sum_{(X_j,Y_j)\in{{D}_{i}^{test}}}[V_{\rm{GAN}}(G_{\omega_G},D_{\omega_D})+w_1V_{L_2}(G_{\omega_G})-w_2V_{\rm{SSIM}}(G_{\omega_G})]$
		\FOR{each sample ${X}^{i}$}
		\STATE Construct corresponding corrected sample $\hat X^i(\omega_{G_i})=G_{\omega_{G_i}}({X}^{i},Z)$ and divide them into $16\times16$ blocks $\left(X^i_{kl},\hat X^i_{kl}(\omega_{G_i})\right)_{16\times16}$
        \STATE Calculate the block SSIM losses for $\left(X^i_{kl},\hat X^i_{kl}(\omega_{G_i})\right)$'s, $L_{\rm{SSIM}}^{1,1,1}\left(X^i_{kl},\hat X^i_{kl}(\omega_{G_i})\right)$ according to equation \eqref{eq: SSIM-def}
        \STATE Select three blocks $\left(X^i_{j},\hat X^i_{(j)}(\omega_{G_i})\right)$ for $j=1,2,3$ with the lowest SSIM values and calculate the differences, $\Delta X^i_{(j)}(\omega_{G_i})=X^i_{(j)}-\hat X^i_{(j)}(\omega_{G_i})$
        \STATE If the lowest block SSIM loss $L_{\rm{SSIM}}^{1,1,1}\left(X^i_{(1)},\hat X^i_{kl}(\omega_{(1)})\right)\geq\mu_0$, then $\bm{1}_{X^i}\{{\rm normal}\}=1$; otherwise, $\bm{1}_{X^i}\{{\rm normal}\}=0$
		\ENDFOR
		\WHILE {not converged (or epoch \textless 300 )}
		\STATE Sample a batch of labeled images for training binary classifier $p_{\theta_i}$
		\STATE Evaluate the cross entropy of the binary classifier $\mathcal{L}_i(\theta_i,\omega) =-\E_{X^i,Z}\biggl[\bm{1}_{X^i}\{\mathrm{normal}\}p_{\theta_i}\left(\Delta X^i_{(1)}(\omega_{(1)}), \Delta X^i_{(2)}(\omega_{(1)}),\Delta X^i_{(3)}(\omega_{(1)})\right)+\left(1-\bm{1}_{X^i}\{\mathrm{normal}\}\right)\left[1-p_{\theta_i}\left(\Delta X^i_{(1)}(\omega_{(1)}), \Delta X^i_{(2)}(\omega_{(1)}),\Delta X^i_{(3)}(\omega_{(1)})\right)\right]\biggl]$
		\STATE Evaluate the gradient of the binary classifier $\bigtriangledown_{\theta_i} \mathcal{L}_i(\theta_i,\omega)$
		\STATE Update $\theta_i$: $\theta_i\leftarrow \theta_i-\gamma \bigtriangledown_{\theta_i} \mathcal{L}_i(\theta_i,\omega)$
		\ENDWHILE
		\STATE Apply the binary classifier to predict other unlabeled blocks of this line $\overline{Y^i} = p_{\theta_i}(X^i,G_{\omega_G}(X^i,Z))$
		\ENDFOR}
	\end{algorithmic} 
\end{algorithm}

\paragraph{Performance evaluation.}
 
The experiment consists of two stages. 
In the first stage, we train our meta-learning framework with
 100 known anomaly samples; and in the second stage, with 50 known anomaly samples. This experiment is to test the accuracy and the robustness of our meta-learning framework.

 We consider five indicators to measure the performance of anomaly detection: the confirmed anomalies, the confirmed anomalies in official anomaly record, the missed anomalies in official anomaly record, the newly-discovered anomalies besides official anomaly record and the percentage of newly-detected anomalies. As a reference, 
\begin{itemize}
 \item {\em Confirmed Anomalies} refers to the number of anomaly samples detected by our meta-learning framework and confirmed by operation staff.
 \item {\em Confirmed Anomalies in Official Anomaly Records} refers to the number of anomaly samples in the Official Anomaly Records that are detected by our meta-learning framework and confirmed by operation staff.
 \item The {\em Missed Anomalies in Official Anomaly Records} refers to the number of anomaly samples in the Official Anomaly Records which are missed by our meta-learning framework.
 \item The {\em Newly-Discovered Anomalies} refers to the number of anomaly samples that are detected by our meta-learning framework and confirmed by operation staff, but were missing in the Official Anomaly Record.
 \item The {\em Percentage of Newly-Discovered Anomalies} refers to the {\em Newly-Discovered Anomalies} divided by the {\em Official Anomaly Record}
\end{itemize}


\begin{table}[!ht]
\caption{Experiment Results with Limited Labeled Samples}
\label{tab: labeled}
\begin{subtable}{1.0\textwidth}
\caption{With 100 Labeled Samples}
\label{subtab: 100}
\begin{tabularx}{1.0\textwidth}{@{}>{\bfseries} p{0.1\textwidth}*{7}{X}@{}}\toprule
  & {\bf Total Number of Images (A)} & {\bf Official Anomaly Record (B)} & {\bf Confirmed Anomalies (C) } & {\bf Confirmed Anomalies in Official Record (D) } & {\bf Missed Anomalies in Official Record (E)} & {\bf Newly Detected Anomalies (F)} & {\bf Percentage of Newly Detected Anomalies (G)} \\
  \toprule
 Test Line 1 &173702&12&83&12&0&71&591\% \\
 Test Line 2 &103302&1&14&1&0&13&1300\% \\
 Test Line 3 &149054&7&56&7&0&49&700\% \\
 Test Line 4 &104250&26&112&26&0&86&330\% \\
 Test Line 5 &168666&34&752&34&0&718&2111\% \\
 \bottomrule
\end{tabularx}
\end{subtable}
\begin{subtable}{1.0\textwidth}
\caption{With 50 Labeled Samples}
\label{subtab: 50}
\begin{tabularx}{1.0\textwidth}{@{}>{\bfseries} p{0.1\textwidth}*{7}{X}@{}}\toprule
  & {\bf Total Number of Images (A)} & {\bf Official Anomaly Record (B)} & {\bf Confirmed Anomalies (C) } & {\bf Confirmed Anomalies in Official Record (D) } & {\bf Missed Anomalies in Official Record (E)} & {\bf Newly Detected Anomalies (F)} & {\bf Percentage of Newly Detected Anomalies (G)} \\
  \toprule
 Test Line 1 &173702&12&74&12&0&62&516\% \\
 Test Line 2 &103302&1&3&1&0&2&200\% \\
 Test Line 3 &149054&7&49&6&1&43&614\% \\
 Test Line 4 &104250&26&61&25&1&36&138\% \\
 Test Line 5 &168666&34&555&34&0&521&1532\% \\
 \bottomrule
\end{tabularx}
\end{subtable}
\end{table}

\begin{table}[!ht]
\caption{ Performance of Deployment in High-Speed Rail System}
\label{tab: unlabeled}
\begin{subtable}{1.0\textwidth}
\caption{Anomaly Detection Ability}
\label{subtab: anomaly}
\begin{tabularx}{1.0\textwidth}{@{}>{\bfseries} p{0.15\textwidth}*{4}{X}@{}}\toprule
  & {\bf Total Number of Images} & {\bf Official Anomaly Record} & {\bf Detected \& Confirmed Anomalies} & {\bf Percentage of Newly Detected Anomalies} \\
  \toprule
 Test Line 1 & 173702 & 12 & 38 & 230\% \\
 Test Line 2 & 103302 & 1 & 4 & 300\% \\
 Test Line 3 & 149054 & 7 & 21 & 250\% \\
 Test Line 4 & 104250 & 26 & 64 & 162\% \\
 Test Line 5 & 68666 & 34 & 414 & 1170\% \\
 \bottomrule
\end{tabularx}
\end{subtable}
\begin{subtable}{1.0\textwidth}
\caption{Workload Reduction}
\label{subtab: workload}
\begin{tabularx}{1.0\textwidth}{@{}>{\bfseries} p{0.10\textwidth}*{3}{X}@{}}\toprule
  & {\bf Total Number of Images} & {\bf Detected Suspected Anomalies} & {\bf Workload Reduction}\\
  \toprule
 Test Line 1 & 173702 & 205 & 99.89\% \\
 Test Line 2 & 103302 & 16 & 99.98\% \\
 Test Line 3 & 149054 & 222 & 99.85\% \\
 Test Line 4 & 104250 & 119 & 99.89\% \\
 Test Line 5 & 68666 & 719 & 99.75\% \\
 \bottomrule
\end{tabularx}
\end{subtable}
\begin{subtable}{1.0\textwidth}
\caption{Time Reduction}
\label{subtab: time}
\begin{tabularx}{1.0\textwidth}{@{}>{\bfseries} p{0.10\textwidth}*{4}{X}@{}}\toprule
  & {\bf Total Number of Images} & {\bf Estimated Time for Human (Hours)} & {\bf Estimated Time for Machine (Hours)} & {\bf {Time Reduction}}\\
  \toprule
 Test Line 1 & 173702 & 1447.5 & 48.3 & 96.7\% \\
 Test Line 2 & 103302 & 860.9 & 28.7 & 96.7\% \\
 Test Line 3 & 149054 & 1242.1 & 41.4 & 96.7\% \\
 Test Line 4 & 104250 & 868.7 & 29.0 & 96.7\% \\
 Test Line 5 & 68666 & 1405.5 & 46.9 & 96.7\% \\
 \bottomrule
\end{tabularx}
\end{subtable}
\end{table}


\paragraph{Results.} 
The test results for these two stage of experiment are shown in Table \ref{subtab: 100} and Table \ref{subtab: 50}. 
\begin{itemize}
\item 
Despite the limited known anomaly samples, our meta-learning framework rarely misses any anomalies that were previously detected by human inspectors, as Column E shows. In fact, after manual inspection, it is confirmed that all anomaly cases missed by our framework are due to internal structural changes which were not captured by images. 

\item Moreover, despite the reduction in the number of labeled samples from 100 to 50, comparing the detected number of anomaly cases between Column D in Table \ref{subtab: 100} and Column D in Table \ref{subtab: 50} for all rail lines shows the robustness of our meta-learning framework for anomaly detection. 

\item In spite of limited known anomaly samples, our meta-learning framework can detect significantly more, in fact several time more anomalies than by human inspectors, as indicated in Column G of Table \ref{subtab: 100} and Table \ref{subtab: 50}.
\end{itemize}

\section{Deployment and Performance}
The robustness and high accuracy of our framework leads to its deployment to the high-speed rail inspection system. Since 2021 this system has been deployed in the following high-speed railways of China: the Beijing--Shanghai line, the Beijing--Harbin line, the Beijing--Guangzhou line, the Lanzhou--Xinjiang line, and the Yichang--Wanzhou line.

The entire inspection system consists of three components: comprehensive track inspection vehicles, the cloud-based storage and processing system, and the client-based operating system. The comprehensive track inspection vehicle is a four-axle inspection vehicle and adopts a dual power transmission system. It uses charged-coupled device cameras to collect high-definition images of the track in real time during vehicle operation and records the corresponding position data. The power of the vehicle is $2\times353$kW, and the running speed is 160 km/h. On average, 1000 images are captured for every kilometer, and the resolution of each picture is about $5120\times5120$. Collected images are sent to the cloud-based storage and processing system in real time through the wireless network. As part of operation trials, our meta-learning framework is deployed in the cloud-based storage and processing system to analyze the collected images in real time.  Suspicious anomalies reported by our system are sent to the client-based operating system, which are then sent to the maintenance personnel for review and double-checking to ensure safety for the train operation.

In the system, a threshold value $\mu_0=0.95$ is adopted to determine that any blocks with SSIM value lower than the threshold are anomalies. Results of the deployment are reported in Table \ref{tab: unlabeled}. 
\begin{itemize} 
\item Our meta-learning framework has managed to detect several times more new anomalies than human inspectors, without prior labeled data. It has learned the difference between foreign objects and background
images, and has detected most foreign objects in each line
(Table \ref{subtab: anomaly}).

\item Our framework has reduced the workload of inspecting millions of images to simply double-checking only hundreds of detected images, over 99.7\% workload reduction (Table \ref{subtab: workload}).

\item  Using our framework, it takes less than two days for anomaly detection, instead of several months of inspection, with over 96\% time saved (Table \ref{subtab: time}).

 \end{itemize}

\section{Conclusion}
In this work, we propose a deep learning framework for anomaly detection: meta-learning   with incorporation of GANs and SSIMs. Our framework has been deployed in the high-speed rail inspection system in China since 2021. Its superb performance demonstrates that our meta-learning framework is promising for general anomaly detection tasks: it is sharp and robust in identifying anomalies and capable of significantly reducing workload.

\bibliographystyle{plain}
\bibliography{meta-learning.bib}
\end{document}